\pdfoutput=1

\documentclass[11pt]{article}

\usepackage[]{acl}

\usepackage{times}
\usepackage{latexsym}

\usepackage[T1]{fontenc}

\usepackage[utf8]{inputenc}

\usepackage{microtype}

\usepackage{graphicx}
\usepackage{amssymb}
\usepackage[ruled, lined, noend]{algorithm2e}

\title{Center-Embedding and Constituency in the Brain\\
and a New Characterization of Context-Free Languages}

\author{Daniel Mitropolsky, Adiba Ejaz, Mirah Shi, Mihalis Yannakakis, Christos H. Papadimitriou \\
Dpeartment of Computer Science \\
Columbia University \\
NewYork, NY 10027}

\begin{document}
\maketitle
\begin{abstract}
A computational system implemented exclusively through the spiking of neurons was recently shown capable of syntax, that is, of carrying out the dependency parsing of simple English sentences.  We address two of the most important questions left open by that work: constituency (the identification of key parts of the sentence such as the verb phrase) and the processing of dependent sentences, especially center-embedded ones. We show that these two aspects of language can also be implemented by neurons and synapses in a way that is compatible with what is known, or widely believed, about the structure and function of the language organ\footnote{Code available \url{https://www.github.com/dmitropolsky/assemblies}.}. Surprisingly, the way we implement center embedding points to a new characterization of context-free languages.
\end{abstract}

\section{Introduction}
How does the brain make language?  Even though it is universally accepted that language is produced through the activity of the brain's molecules, neurons, and synapses, there has been extremely slow progress over the past decades in the quest for pinpointing the {\em neural basis of language,} that is, the precise biological structures and processes which result in the generation and comprehension of language --- see \citet{Friederici} for an excellent overview of a major direction in the theory of the language organ.  In a recent advance in this direction, a {\em parser of English} was implemented \cite{Parser} in the computational system known as the Assembly Calculus (AC) \cite{PNAS}, a biologically plausible computational framework for implementing cognitive functions.  The basic data structure of the AC is the {\em assembly of neurons}, a large set of neurons representing an idea, object, episode, word, etc. --- a brief description of the AC and its brain-like execution environment is given in Section \ref{ACoverview}.  The Parser is implemented through assembly operations, and thus ultimately by the actual spiking of stylized neurons.  Its input is a sequence of words, and in response to it the Parser produces, as a detectable substructure of the neural circuit, a correct {\em dependency parse} of the sentence. For this to happen, neural representations of the words are assumed to be already in place in a brain area called {\sc Lex} (for {\em lexicon}\/). For each new word, all neurons in the corresponding representation spike, and the representations contain enough grammatical information to cause a cascade of neural activity which results in the correct parsing of the sentence.  It was shown through experiments in \citet{Parser} that several simple classes of sentences can be parsed this way.  It is important to remember that the Parser in \citet{Parser} works exclusively through the spikes of biologically plausible neurons, and hence it can be seen as a proof of concept --- a concrete hypothesis even --- about the way syntactic analysis of language happens in the brain.

Several research directions were left open in \citet{Parser}, and preeminent among them were these two: (a) how can the parser be extended so that dependent clauses, and especially center-embedded ones, are parsed correctly? and (b) how could the {\em constituency} representation of the sentence (the tree with "Sentence" and "Verb Phrase" as its internal nodes and the ``Subject,'' ``Verb,'' and ``Object'' as leaves) be produced --- there is experimental evidence that the main constituents of a sentence are indeed created in Broca's area during sentence processing (e.g. \citet{ZF}). 

Parsing center-embedded sentences presents a serious conceptual difficulty: the parsing of the embedding sentence must be interrupted, and then be continued after the embedded sentence is parsed.  This is the nature and essence of recursion.  A mechanism for recovering the state of the parser at the moment of the interruption seems thus necessary.  It would be tempting to posit that recursion in the brain happens as it does in software, through the creation of a stack of activation records, but as we shall discuss this is not biologically plausible.

In this paper we pursue these two unresolved research goals and make significant progress on both. We start our description with constituents, which is the simpler narrative. The Parser in \citet{Parser} consists of the brain area {\sc Lex} where the lexicon resides (believed to be a part of the medial temporal lobe, MTL), as well as several other areas labeled {\sc Subj}, {\sc Verb}, {\sc Obj}, {\sc Det} etc.~corresponding to syntactic roles, believed to be subareas of Wernicke's area; this is where the dependency parse is created.  We show that this architecture can be augmented by new areas and fibers in such a way that the basic constituency tree of the sentence (the three-leaf tree that has inner nodes Sentence and Verb Phrase) can be also built ``on the side'' while the sentence is parsed.  This entails two new brain areas denoted S for sentence and VP for verb phrase. The brain areas in our model are intended to correspond to two well known subareas of Broca's area, BA 43 and BA 44, respectively, where such activity is thought to take place \cite{Friederici}. 

Coming now to the problem of dependent clauses, to handle clauses that are not embedded (say``if dogs are angry, they chase cats''), the  Parser needs only one extra brain area beyond the areas needed by the original \cite{Parser}, namely an area labeled {\sc DS} for ``dependent segment.'' 

To see the problem with embedded sentences, consider the variant ``dogs, when they are angry, chase cats.''  The Parser would recognize ``dogs'' as the subject of the sentence and project the assembly for ``dogs'' to the SUBJ area and change the state as appropriate (inhibit a fiber).  Upon encountering the first comma, the parsing of the sentence is curtailed, and the parsing of a new clause will be initiated. This is not a problem, since brain areas can contain many assemblies at the same time, and the two verbs, subjects, etc.~will not interfere with each other. However, upon the second comma (for simplicity we assume that these clues are always available-- in speech, they may, for instance, by indicated using prosodic or pausal cues), the Parser needs to continue the parsing of the outer clause (that is, the main sentence), and for this {\em it needs to restore the state at the moment the parsing was interrupted by the embedding sentence.}  We could have the state saved in a pushdown store and retrieve it at this point, but we can see no biologically plausible implementation of either Parser state records or a pushdown store.  As we shall point out, the Parser so far is a finite automaton, and thus it is not surprising that it has trouble handling embedded clauses.  

The question is, {\em which biologically plausible departure from finite state machines, and from the model in \citet{Parser}, can handle embedding?}  
We propose the following: The Parser stores the part of the utterance already parsed in some {\em working memory} (as in \citet{neuro}, for example).  When the embedded clause has been parsed, the Parser returns to the beginning of the outer sentence in the working memory and reprocesses it (up to the comma) to restore the state. We only need to execute the action sets of the words in the first part of the sentence, an activity that we we call {\em touching,} which is, in terms of elapsed time, an order of magnitude faster than parsing the first time. Once the first comma is seen, the embedded clause is skipped (without state changes) until the second comma, and parsing of the outer sentence is resumed from the recovered state. Any embedded clause can be handled with the touching maneuver --- non-embedded dependent clauses are much simpler.

We show by processing several examples that several forms of dependent and embedded clauses can be correctly parsed (that is, a correct dependency graph of the whole sentence can be retrieved from the device after processing). If the embedded clause has its own embedded clause, the trick can be repeated.

Very surprisingly to us, this extension of the Parser, arrived at strictly through considerations of biological plausibility and ease of implementation, yields a rather unexpected theorem in formal language theory: if one defines an extension of nondeterministic finite-state automata to capture the operation of the enhanced Parser, with the extra capabilities of (a) marking the current input symbol and (b) reverting from the current input symbol to the previously marked symbol that is closest to the current one, then the class of languages accepted by these devices --- call them {\em fallback automata} --- coincides with the context-free languages!

\section{Background}
\subsection{The Assemblies Model} \label{ACoverview}
How does the brain beget the mind?  How do molecules, neurons, and synapses effect reasoning, planning, emotions, language?  Despite tremendous progress in the two extremes of scale --- cognitive science and neuroscience --- we do not know how to bridge the scales.  According to Nobel laureate Richard Axel \cite{Axel}, ``We don't have a Logic for the transformation of neuronal activity to thought and action.  I consider discerning (this) Logic as the foremost research direction in neuroscience.''  Notice the use of the word ``Logic'' whereby a distinguished experimentalist dreams of a formal system...

Recently, a computational framework called the Assembly Calculus (AC) was proposed whose precise intention is to be this Logic: to model the brain at the level of cognitive function through the control of a {\em dynamical system of spiking neurons.} This section describes the variant of the AC used in \citet{Parser} and the present paper. 

We start with a mathematical model of the brain: a finite number $a$ of brain {\em areas} $A, B,...$ each containing $n$ excitatory neurons.  Every two neurons $i,j$ in each area have a probability $p$ of being connected by a synapse.  Each synapse $(i,j)$ has a nonnegative weight $w_{ij}>0$, initially $1$, say, which changes dynamically.  For certain {\em unordered} pairs of areas $(A,B), A\neq B$, there is a {\em fiber} connecting them, a random directed {\em bipartite} graph connecting neurons in $A$ to neurons in $B$ and back, again with probability $p$ independently for each possible synapse. Thus, the brain is a large directed graph with $an$ nodes and random weighted edges.  

Time is discrete (in the brain each time step is throught to be about 20 ms). The state of the dynamical system at time $t$ has two components: the weights of the synapses $w_{ij}^t$, and the set of neurons that {\em fire} at time $t$.  That is, for each neuron $i$ we have a state variable $f_i^t\in \{0,1\}$ denoting whether or not $i$ {fires} at time $t$.  The state transition from time $t$ to time $t+1$ is computed thus:
\begin{enumerate}
    \item For each neuron $i$ compute its {\em synaptic input} $SI_{i}^t=\sum_{(j,i)\in E, f_j^t=1}w_{ji}^t$, that is, the sum total of all weights from pre-synaptic neurons that fired at time $t$.
    \item For each neuron $f_i^{t+1}=1$ --- that is, $i$ fires at time $t+1$ --- if $i$ is among the $k$ neurons {\em in its area} with the highest $SI_i^t$ (breaking any ties arbitrarily).
    \item For each synapse $(i,j)\in E$,\\
    $w_{ij}^{t+1}=w_{ij}^t(1+ f^t_i f^{t+1}_j \beta)$; that is, a synaptic weight increases by a factor of $1+\beta$ if and only if the post-synaptic neuron fires at time $t+1$ and the pre-synaptic neuron had fired at time $t$.
\end{enumerate}

These are the equations of the dynamical system.  The AC also has commands for the  high-level {\em control} of the system. A fiber can be {\em inhibited} (that is, prevented from carrying synaptic input to other areas) and {\em disinhibited} (inhibition is canceled). Also, a set $x$ of $k$ neurons in an area can be made to fire by the command fire($x$) (this is most relevant in connection to {\em assemblies,} defined next.


The {\em state} of the system contains the firing state of each neuron, the edge weights $w_{ij}$, and inhibition information. 

{\em Assemblies of neurons} are a critical emergent property of the system. An {\em assembly} is a special set of $k$ neurons, all in the same area, that are {\em densely interconnected} --- that is, these $k$ neurons have far more synapses between them than random, and these synapses have very high weights.  This renders assemblies {\em stable representations} for representing in the brain objects, words, ideas, etc. 


How do assemblies emerge?  Suppose that at time $0$, when nothing else fires, we execute fire$(x)$ for a fixed subset of $k$ neurons $x$ in area $A$ (these $k$ neurons will always correspond to a previously created assembly\footnote{Initially, assemblies are created by projection from stimuli from the outside world coded in the sensory cortex.}), and suppose that there is an adjacent area $B$ (connected to $A$ through a disinhibited fiber) where no neurons currently fire.  Since assembly $x$ in area $A$ fires at times $0,1,2,\ldots$ (and ignoring all other areas), it will effect at times $1,2,\ldots$ the firing of an evolving set of $k$ neurons in $B$, call these sets $y^{1},y^{2},\ldots$.  It is shown in \citet{PNAS} that, with high probability (where the probability space is the random connectivity of the system), the sequence $\{y^t\}$  eventually converges to a stable assembly $y$ in $B$, called {\em the projection of $x$ in $B$}. The new assembly will be strongly interconnected, and also has strong connections from with $x$: If one of the two assemblies henceforth fires, the other will follow suit.


In \citet{Parser} this operation was generalized for the paper's purposes. Suppose an assembly $x$ in area $A$ fires repeatedly and there are many areas downstream --- not just a single area $B$ as before --- and these areas are connected by disinhibited fibers in a way that forms a {\em tree.}  Then a sequence of project operations, denoted project$^*$, creates a tree of assemblies, with strong synaptic connections between them.
In \cite{Parser}, it is these synaptic connections between projected word assemblies that constitute the valid dependency parse tree created by the Parser.

\subsection{The Parser}
The basic architecture of the Parser in \citet{Parser} consists of several brain areas connected by fibers. {\sc Lex} special, and contains representations of all words. Upon input of a new word (in brain reality, read or heard) the corresponding representation is excited, and upon firing it executes the {\em action set} of the word, commands which capture the grammatical role of the word.  These commands open and close (disinhibit and inhibit) certain fibers. Then the project* operation is executed: all active assemblies fire.  By this scheme, it was shown in \citet{Parser} that many categories of English sentences (and Russian as well) can be parsed correctly (the correctness can be verified because the running of the parser on an input sentence leaves a retrievable graph structure within and between brain areas, which constitute a correct dependency graph of the sentence).
This completes the description of the Parser's architecture.  We provide a figure with an example sentence parsed, and refer the interested reader to \citet{Parser}.

\subsection{Neuroscience} 
We believe that our Parser represents a reasonable hypothesis for parsing in the brain. The neurobiological underpinnings of Assembly Calculus, with which our model is built, and the original Parser of \citet{Parser}, which ours extends, are presented more fully in that paper. Briefly, Assembly Calculus is based on established tenets of neuron biology, including that neurons fire when they receive sufficient excitatory input from other neurons, the atomic nature of neuron firing, and a simplified narrative of synaptic Hebbian plasticity (see for instance \citet{NeuralScience}, Chapters 7, 8, and 67). Assemblies, in turn, are an increasingly popular hypothesis for the main unit of higher-level cognition in modern neuroscience-- first hypothesized decades ago by Hebb, they have been identified experimentally \cite{Harris} displacing previously dominant theories of information encoding in the brain, see e.g.~\citet{Eichenbaum}. With what regards the higher-level Parser architecture, language processing appears to start with access to a {\em lexicon}, a look-up table of word representations thought to reside in the left medial temporal lobe (MTL), motivating the inclusion of an area {\sc Lex}. After word look-up, activity in the STG is thought to signify the identification of syntactic roles. Overall, the Parser generates a hierarchical dependancy-based structure that from a sentence that is processed incrementally, which we believe models something like the creation of hierarchical structures in Broca's areas in experiments such as \citet{Poeppel}.

\section{Constituency}

Constituency parsing revolves around the idea that words may lump into a single assembly. The noun subject of a sentence --- along with its dependent adjective(s) and determinant(s) if any --- form the ``Subject," while the verb and object form the ``Verb Phrase." At the coarsest level, the ``Subject" and ``Verb Phrase" then form the ``Sentence." We modify the underlying framework of the Parser to assign this syntactic structure to a sentence. 

We add two new brain areas that hold the ``Verb Phrase" and ``Sentence," {\sc VP} and {\sc S} respectively, as well as fibers between {\sc Verb} and {\sc VP}, {\sc Obj} and {\sc VP}, {\sc Subj} and {\sc S}, and {\sc VP} and {\sc S}. These fibers remain disinhibited throughout parsing so that the constituency tree is built concurrently as we parse the sentence. That is, when the verb is processed, a corresponding assembly is formed in {\sc VP}, and upon encountering the object, the assemblies in {\sc Verb}, {\sc Obj}, and {\sc VP} fire together to form in {\sc VP} what is now the merge of assemblies representing the verb and object. Parallel to this process, assemblies fire along the {\sc Subj} to {\sc S} and {\sc VP} to {\sc S} fibers so that the final assembly in S represents the joining of the ``Subject" and ``Verb Phrase." 

\paragraph{Experiments.}
We extend the implementation of the Parser in Python to incorporate these new abilities. Additionally, we tailor the readout algorithm (which in \citet{Parser} recovers the dependency tree) to output the desired tree rooted in {\sc S}. To verify that the Parser produces the correct constituency tree, we provide a test set of 40 sentences constructed from 20 syntactic patterns that include variations in word orderings; additions of determinants, adjectives, adverbs, and prepositional phrases; and both transitive and intransitive verbs. The Parser generates the correct constituency trees on all of our given test cases. Importantly, the constituency Parser can handle any sentence structure that the original Parser can handle. 

As in the original parser, we execute 20 firing epochs of project$^*$ to allow the dynamical system to stabilise. The multiple concurrent projections into S and VP cause a slowdown by a factor of 2.5 relative to the dependency parser, resulting in a frequency of 0.5-1.3 seconds/word. 

\section{Embedded Sentences}
To handle embedded sentences, the Parser requires a new area {\sc DS} (for dependent segment) to handle dependent and embedded clauses.  Additionally, we need modifications that recover the state {\em before} an embedded clause (which we recover when we finish parsing an embedded clause, i.e., upon a ``right comma").

In particular, we assume that there is a {\em working memory area} which holds the words which have already been processed, in sequence. We assume for simplicity that the parser can always recognize the beginning of a dependent sentence --- that this is always possible through simple clues such as a comma (in text), prosodic cues (in speech), and/or a {\em complementizing} pronoun or preposition, such as ``who'', ``if'', or ``that". We also assume that there is a unique sentence or clause at each depth, though with minor modifications we can handle the more general case.


To handle center-embedded clauses, the parser utilizes a limited working memory: it must remember the sentence up to the point when an embedded clause begins in order to efficiently reprocess these words after parsing the embedded clause. More concretely, when a center embedding is detected, the Parser ``cleans the slate"; that is, the Parser state (the fiber and area states) is reinitialized in order to parse the new embedded clause, which is parsed normally until its end is detected. At this point the Parser has to restore its last state when it was parsing the outer clause. To do so, it reinitializes the state again and reprocesses the sentence from the beginning of the outer clause up to the interruption. However, this re-processing is special: the parser only ``touches'' the words, by which we mean that for each word we apply its action sets (inhibiting or disinhibiting areas/fibers), fire the entire system exactly once to reactivate existing assemblies that were formed in the initial parse of the fragment (when we initially parsed this part of the sentence, we used project*, which fires the entire system 20+ times in order for assemblies to form and converge). In this way, touching is by an order of magnitude faster than parsing on initial input of the word --- both in our simulation and in the hypothesized language organ --- since it is only recovering the {\em preexisting} structure.  Note that when the first comma is scanned, the parsing of the outer sentenced is resumed from the second comma, skipping the (already parsed) embedded clause.

The remaining difficulty is in linking the outer and inner clauses. The last word of the outer clause before the inner sentence functions as a ``signature" for the outer clause. This signature may reside in SUBJ or OBJ, for instance. After parsing the fragment of the outer clause pre-interruption, we project from the relevant signature area to {\sc DS}. Then, on the verb of the inner clause, we project from both {\sc LEX} and {\sc DS} into {\sc VERB} simultaneously. This way, we can later recover the root verb of the inner clause via the signature assembly of the outer sentence (through projection into {\sc DS} and subsequently {\sc Verb}). to {\sc DS}, and then to {\sc VERB}.


\paragraph{Experiments.} We extend the implementation of the dependency Parser in Python to incorporate touching, linkage, and recursion within the main Parser loop. In principle, the developed Parser can handle arbitrarily many levels of embedding. We test on 20 sentences sampled from 5 different embedding structures, with depths 0, 1, and 2 (informed by the lack of three or greater depth sentences in ordinary language). Our test cases feature center-embedding, edge-embedding, mixed embedding, and relative clauses modifying the subject or the object. We assume there is a unique clause at each depth. The Parser generates the expected dependency tree on all of our given test cases. 

Again, we execute 20 firing epochs of project$^*$. The modified parser preserves the speed of the original, with a negligible increase in time for linkage projections and touching.  Despite several challenges created by the added complexity of sentence embedding, the linkages between projected assemblies correspond to the correct dependency graph in all sentences.

\begin{algorithm}
\SetKwInOut{Input}{input}
\SetKwInOut{Output}{output}
\Input{a sentence $s$, depth $d$}
\Output{representation of dependency parse of $s$, rooted in {\sc Verb}}
\SetKwFunction{Fparse}{{\normalfont \em parse}}
\SetKwProg{Fn}{Function}{:}{}
\Fn{\Fparse{\upshape $s$, $d\gets 0$}}{
    \ForEach{\upshape word $w$ in $s$}{
        \If{\upshape $(d+1)$-depth clause begins after $w$}{
            disinhibit({\sc DS}) \;
            disinhibit({\sc DS}, {\sc Area}($w$)) \;
            {\em project}* \;
            inhibit({\sc DS}, {\sc Area}($w$)) \;
            inhibit({\sc DS}) \;
        }
        \ElseIf{\upshape $w$ begins (d+1) depth clause}{
             $d\gets d+1$ \;
             clear the slate \;
        }
        \ElseIf{\upshape $w$ ends embedded sentence}{
            $d\gets d-1$ \;
            \ForEach{\upshape word $y$ before $w$}{
                activate $y$ in \sc{LEX} \;
                {\em fire} disinhibited areas \;
            } 
        }
        \If{\upshape $d>0$, {\sc Area}($w$) = {\sc VERB}}{
            disinhibit({\sc DS}) \;
            disinhibit({\sc DS}, {\sc VERB}) \;
        }
        execute $w$ actions and {\em project}* \;
        inhibit({\sc DS}) \;
        inhibit({\sc DS}, {\sc VERB}) \;
    }
}
\BlankLine
 \caption{Enhanced Parser, main loop}
\end{algorithm}

\begin{figure*} 
  \includegraphics[scale=0.8, trim=15 70 0 0, clip]{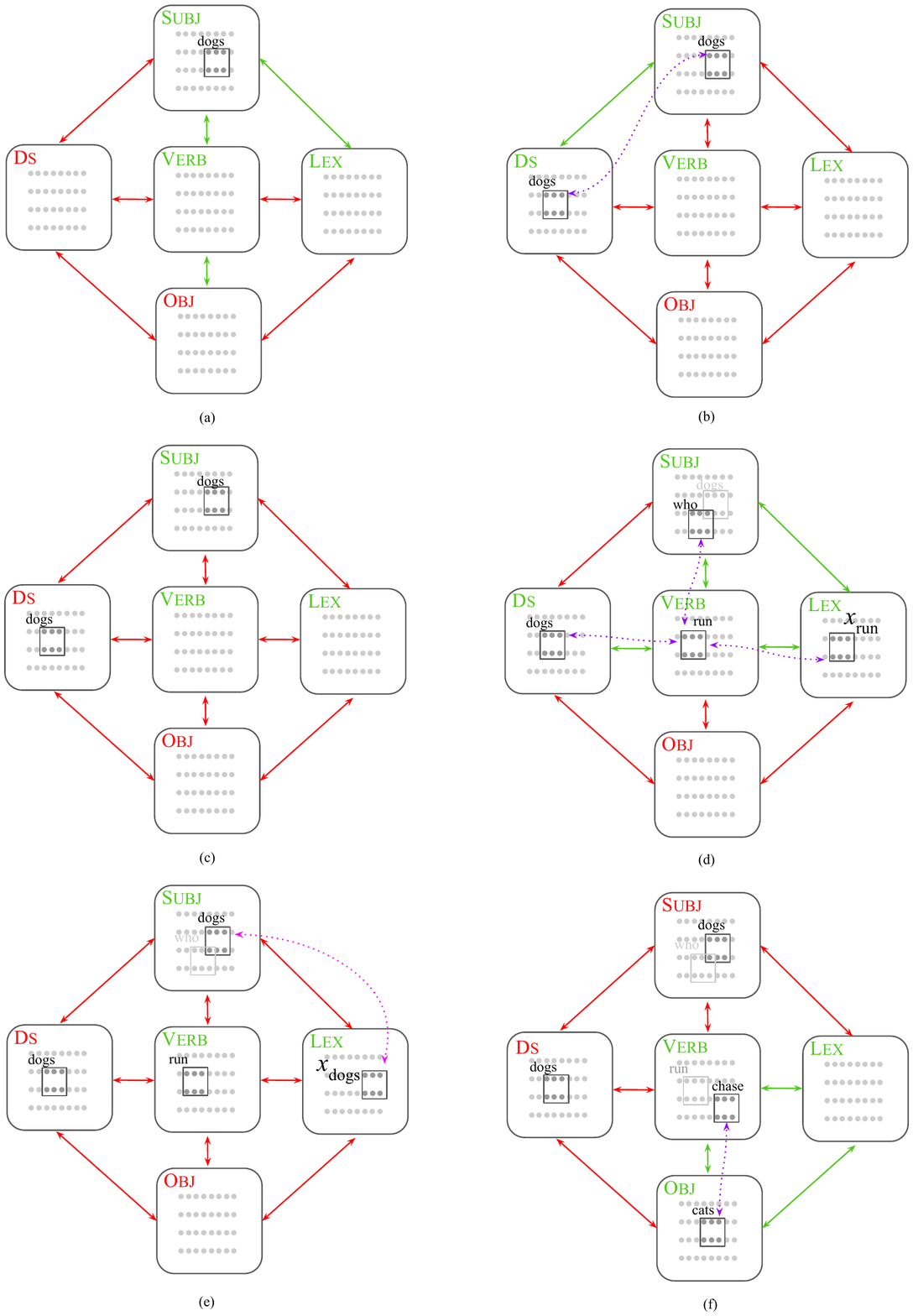}
  \caption{Snapshots of the Parser while processing the center-embedded sentence ``\textit{dogs, when they run, chase cats}." Green arrows represent fibers that have been disinhibited between each stage; red arrows represent inhibited fibers. Purple dotted arrows indicate assemblies that have been connected through project$^*$. Pink dotted arrows indicate assemblies touched through activating their corresponding assemblies in LEX. (a) Outer sentence, up until beginning of inner sentence, parsed; (b) Assemblies in the signature and DS areas linked; (c) Fiber and area states reset in anticipation of a new sentence; (d) Inner sentence parsed and linked to outer sentence through the DS area; (e) Outer sentence revisited: each word prior to dependent sentence is touched, restoring Parser state of stage (a); (f) Remainder of outer sentence parsed.}
  \label{fig:ex}
\end{figure*}

\section{A Little Formal Language Theory}
The Parser, without the embedded sentence module, is a {\em finite state device.}  The reason is that the Parser's state is an acyclic subgraph of the graph of brain area and fibers (excluding LEX).  It is no wonder then that sentence embedding (intuitively, a feature involving recursion and therefore moving us out of the realm of regular languages) requires an extension, and among many other options that seemed to us less biologically plausible we chose to revisit the outer sentence and restore the state of the Parser.  It is now natural to ask:  Can the operation of the Parser when handling embedded sentences be seen as a more powerful genre of automaton? This is the motivation for the results in this section.

\medskip\noindent {\bf Definition:}  A {\em fallback automaton (FBA)} is a tuple $A=(\Sigma,K,I,F,\Delta),$ where (as in a nondeterministic finite-state automaton) $\Sigma$ is a nonempty finite set of symbols, $K$ is a set of states, $I$ is the set of initial states, $F$ is the set of accepting states, and $\Delta$ is the transition relation. 

Define the {\em type set} $T = \{f,s\}\cup K$. Whereas in nondeterministic FSAs $\Delta \subseteq (\Sigma\times K)\times K$, the transition relation of FBAs is more complex.

$$\Delta\subseteq
((\Sigma\times T \times K)\times(K \times \{s,\checkmark,\leftarrow\})$$

The automaton is {\em nondeterministic,} reflecting the nondeterministic nature of parsing in general, due to ambiguity and polysemy.  The transition relation  can be understood thus: symbols on the tape are marked by a type, either $f$ (for {\em fresh}), $s$ (for {\em seen}), or by a state $q \in K$. Initially, all symbols are fresh and hence marked $f$. When a symbol is marked $s$ or by a state of $K$, it is not being scanned for the first time --- the automaton may scan a symbol multiple times. The first time a symbol is scanned, its type is changed from $f$ to $s$, or, if the rule outputs $\checkmark$, it is set to the current state $q$. Subsequently, after a symbol marked by a state is scanned, the type reverts to $s$.

To formalize the operation of the automaton, at each step, there is a {\em tape} $x\in (\Sigma\times T)^*$; we denote its $i$th symbol as $x_i=(\sigma_i,t_i)$,
where $\sigma_i\in\Sigma$ and $t \in T$.  The automaton is in a state $q\in K$ and the $i$th symbol $x_i$ is scanned.  The overall configuration is thus $(x,q,i)$.  In the initial configuration, the type of every symbol is $f$, $q\in I$ and $i=1$.  

If the next configuration after $(x,q,i)$ is $(y,r,j)$ then:
\begin{itemize}
    \item $y$ is identical to $x$ except that the type of the $i$th symbol may have changed: $f$ must become $s$ or the machine's current state $q$, $s$ stays $s$, and $q \in K$ always become $s$.
    \item unless this is a fallback step, $j=i+1$.
    \item {\em Fallback.}  When scanning a fresh symbol, the automaton may return to a position $j<i$, where $j$ is the largest position $j$ that was marked --- that is, $t_j \in K$.
    Notice that in the next step, $t_j=s$--- the symbol is unmarked.
    \item When scanning a symbol with type $q \in K$, i.e., a symbol fallen back to, the transition can map to only $s$ or $\leftarrow$-- that is, it can fallback again, but it cannot mark the symbol again for fallback.
    \item In all cases, the corresponding pair $((x_i,q)(r,\sigma))$ must be in $\Delta$. $\sigma=s$ means that the $i$-th symbol's type is changed to $s$, $\sigma=\checkmark$ means it is changed to the state $q$, and $\sigma=\leftarrow$ means the step is a fallback.
\end{itemize}
This concludes the definition of the FBA.
We say that a string $x$ in $\Sigma^*$ is accepted by FBA $A$ if there is a sequence of legal steps from a configuration with state in $I$ and tape $x$ with all symbols fresh to a configuration in which the state is in $F$. 

Note that when the FBA falls back to a previous tape location $j < i$, it then passes again over the seen symbols (marked $s$) between $x_j$ and $x_i$, and may do meaningful computation upon this revisiting. Furthermore, an FBA can fallback repeatedly, immediately after a fallback move. Our Parser more closely corresponds to an FBA without this abilities. Hence it is interesting to define a weaker model without this ability:

A {\em weak-FBA} is an FBA with the additional requirements that 1) for any symbol marked $s$, transitions cannot change the state (that is, for all $\alpha \in \Sigma,~q\in K$ and $(\alpha,s,q)\times(q',s) \in \Delta$, we require $q' = q$, in effect, skipping over all $s$ symbols) and 2) for any symbol marked with $q \in K$, the transition must output $(q',s)$ (that is, there are no repeated fallbacks). 

It may seem that FBAs can do more than weak-FBAs. Consider the following language over $\Sigma = \{0,1,\alpha,\beta\}$, $L = \{\alpha^n x \beta^n~:~x \in \{0,1\}^*\}$. By marking every $\alpha$ and falling back on every $\beta$, an FBA can read through $x$ at least $n$ times, a linear dependence. However, a weak-FBA can read each symbol in the tape exactly 1 or 2 times.

Perhaps surprisingly, it turns out that the ability to do additional computation on revisited symbols offers no additional power. More importantly, it turns out that both models recognize a fundamental class of formal language theory. Denote the class of languages accepted by FBAs as {\bf FBA}, that by weak-FBAs as {\bf weak-FBA}, and the class of context-free languages as {\bf CFL}.
We can prove the following:

\medskip\noindent{\bf Theorem:} {\bf weak-FBA} = {\bf FBA} = {\bf CFL}

\medskip\noindent {\bf Proof outline:} To show that {\bf weak-FBA} $\supseteq$ {\bf CFL}, we recall the classic theorem of Chomsky and Sch\"utzenberger \cite{chomsky_shutz} stating that any context-free language $L$ can be written as $L= R \cap h(D_k)$, where $R$ is a regular language, $D_k$ denotes the Dyck language of balanced parentheses of $k$ kinds, and $h$ is a homomorphism, mapping any symbol in the alphabet of $D_k$ to a string in another alphabet. Let us take a context-free language in this form.  Note that FBAs are ideal for accepting $h(D_k)$. The machine uses non-determinism to guess which symbol is represented by the next sequence of characters. When it guesses that it will see the image of a left parenthesis, say `\{', it checks each symbol of $h(\{)$ (and rejects if the sequence of symbols is not $h(\{)$), and marks the final character (with the state $q_{\{})$). For a right parenthesis, after checking for the sequence $h(\})$, it falls back and checks that the symbol fallen back to is marked $q_{\{})$. Intersection with the regular language $R$ is done by simultaneously maintaining the state of the automaton accepting $R$ in a separate component of the FBA's state (in fact, one can show that the languages accepted by FBAs are closed under intersection with regular languages).

To show that {\bf FBA} $\subseteq$ {\bf CFL}, we emulate the execution of a FBA with a {\em push-down automaton} (that is, a non-deterministic finite state automaton with the additional computational power of one stack). By the classic result proved independently by \citet{ChomskyCFG, SchutzenbergerCFG, Evey}, the languages recognized by push-down automata are exactly {\bf CFL}. The emulation uses the following trick: the stack is composed of {\em vectors} of states of length $|K|$. These vectors keep track of the execution of the FBA on every possible state on the sequence of symbols between consecutive pairs of marked symbols, and between the most recent marked symbol and the head. Whenever the FBA falls back and is in state $q$ where $q$ corresponds to the $i$-th coordinate in the stack vectors, the emulation pops the top vector on the stack and jumps to the state in the $i$-th coordinate, as this would be the resulting state {\em had the machine re-read the seen symbols starting in state $q$}. The full proof is technical, and is given in the Appendix.

\section{Discussion} 
The Parser in \citet{Parser} can be seen as a concrete hypothesis about the nature, structure, and operation of the language organ.  Here we elaborate on this hypothesis: First, rough constituency parsing (the creation of the two highest layers of the syntactic --- or constituency --- tree of the sentence) can be carried out simultaneously with the main dependency parsing.  Second, dependent sentences can also be parsed.  For center-embedded sentences, a significant extension of the Parser is required:  A working memory area stores the whole utterance, and the parser returns to the beginning of the utterance to recover the state of the Parser after processing the outer sentence, and and then skips the embedded sentence and continues parsing the outer one.  

Even though this maneuver was motivated by biological realism and programming necessity, we showed that it transforms the device from one that handles only regular languages to one capable of accepting {\em all context free languages} --- and {\em just} these. We find this quite surprising, and possibly significant for the history of linguistic theory: Seven decades ago, Noam Chomsky sought to formalize human language and in the mid 1950s introduced CFLs expressly for this purpose.  In the following two decades, this choice was criticised as too generous (not all features of CFLs are needed) and also as too restrictive (some aspects of natural language are not covered by CFLs). Arguably, this criticism was accepted by Chomsky's school of thought: Grammar remained important, of course, but context-free rules besides $S\rightarrow NP VP$ (right-hand side unordered) were not used often. Much of NLP centered around the dependency formulation of syntax.   Two-thirds of a century later, computer scientists speculating about syntax in the brain came up with a computational trick in order to handle center recursion.  And this maneuver, when formalized properly, leads to a device that can recognize all CFLs.  

Besides speculating on the meaning of this theoretical result, our work suggests a major open problem: If we assume that syntax in the brain is handled in a way similar to the one suggested by the Parser\footnote{A far-fetched assumption, of course, but one that is somewhat justified by the fact that there is no competing theory that we are aware of.}, and all humans are born with a system of brain areas and fibers in their left hemisphere capable of such operation, {\em how do babies learn to use this device?} How are words learned and projected, presumably from the hippocampus, where they are associated with world objects and episodes, to the LEX in the medial temporal lobe?  And how is each of them attached to the correct system of interneurons that are capable of changing the inhibited/disinhibited status of fibers and possibly of brain areas?    

\section{Appendix}
Here we give the full proof of our main theoretical result:

\medskip\noindent{\bf Theorem:} {\bf weak-FBA}$=${\bf FBA}$=${\bf CFL}.

\medskip\noindent{\bf Note:} We say a symbol on the tape is ``marked' whenever its type is some $q \in K$.

\medskip\noindent{\bf Lemma:} For any strong-FBA, there exists a strong-FBA recognizing the same language that is deterministic whenever the input symbol is marked $s$ (that is, for every state and symbol pair $q,\alpha$, there is at most one rule $((\alpha,s,q),(q',s)) \in \Delta$.

\medskip\noindent {\bf Proof:} this is shown using essentially the same reduction from non-deterministic to deterministic finite state automata (FSA), since when at a seen symbol, the FBA cannot mark or fallback and is hence in a FSA-like regime. Concretely, if $K$ is the original state set, the state set of the new FBA is $2^K$. $\Delta$ contains the same rules when the input is in state $f$ or is marked (we represent states of the original FBA with the singleton of that state in $2^K$)-- whenever the input is in state $s$, it transitions to the $\epsilon$-closure of the subset represented by the state (i.e., $(\alpha,s,S)\rightarrow (S',s)$ iff $S'$ is the set of all states that can be reached from a state of $S$ on symbol $\alpha$, before or after any epsilon transitions). Additionally, whenever the FBA reads symbol marked with anything other than $s$, or reaches the end of the tape, if the current state-set $S$ is a non-singleton, it non-deterministically transitions to the singleton of any $q \in S$ (thereby returning to a ``regular" state of FBA, and moving all the non-determinism away from $s$ symbols). The new FBA is determinstic on $s$-inputs and recognizes the same language: any transition through a sequence of $s$-states corresponds to a specific non-deterministic transition to a single state from the final state-set at the first non$-s$ symbol. $\square$

The main technical result is showing that a pushdown automaton (PDA) can simulate a {\bf strong-FBA}, i.e., that {\bf strong-FBA}~$\subseteq$~{\bf CFL}. 

\medskip\noindent {\bf Proof of theorem:} By the lemma, without loss of generality we can assume that the strong-FBA is deterministic when the input has type $s$. Let $K=\{q_1,\ldots,q_t\}$ be the state set of the strong-FBA. The PDA will have state set $K$, and stack alphabet $\Sigma \times K \times K^{|K|}$, that is, tuples of a symbol, a state, and {\em state vectors}. We define the ``$s$-transition on $\alpha$" of a state $q$ to mean the (deterministic) FBA transition rule with left-side $(\alpha,s,q)$.

The PDA simulates the execution of the strong-FBA. When in state $q'$ and on fresh tape symbol $\alpha$, the PDA will:

1) Non-deterministically select a rule with left side $(\alpha,f,q')$ on the left-hand side. Let $(q,\sigma)$ be the right-hand side. That is, $\sigma \in \{s,\checkmark,\leftarrow\}$.

2) Update the state to $q$.

3) If the stack is non-empty, pop the top element $(\beta,p, (r_1,\ldots,r_t))$ from the stack. For each vector coordinate $i \in [t]$, apply the $s$-transition on $\alpha$ to each $r_i$-- push the updated tuple $(\beta,p, (r_1',\ldots,r_t'))$ back to the stack.

4) if $\sigma = \checkmark$, push $(\alpha, q, (q_1,\ldots,q_t))$ to the stack.

5) if $\sigma = \leftarrow$, the corresponding PDA transition does nothing else, but enters the following loop:

5.1) Pop the top element $(\beta, p, (r_1,\ldots,r_t))$ of the stack and sample a rule of the FBA that transitions on the marked symbol, that is, a rule of the form $(\beta,p,q)\rightarrow (q',\sigma)$. Note that $\sigma \in \{s,\leftarrow\}$. If the stack is non-empty, pop the next element, $(\gamma, w, (u_1,\ldots,u_t))$ and ``apply" $(r_1,\ldots,r_t)$ to the state vector-- that is, for each $i \in [t]$, if $u_i = q_j$, replace $u_i$ with $u_i' := r_j$. Push the resulting pair $(\gamma, w, (u_1',\ldots,u_t'))$ back onto the stack.

5.2) if $\sigma = \leftarrow$ (which we call a ``fallback-again" rule of the FBA), the PDA updates the state to $q'$, and returns to the beginning of 5.1). If $\sigma = s$ (which we call a non-``fallback-again" rule), the PDA updates the state to $r_i$ (that is, the $i$-th coordinate in the stack-vector from 5.1), where $i$ is the index of $q'$, i.e. $q'=q_i$. Because the FBA can have ``fallback-again" rules, the PDA can repeat 5.1 multiple times, and stops when it selects a non-``fallback-again" rule (or rejects).

For any execution of the PDA, it will have $|x|+F+M$ transitions, or steps, where $x$ is the input, $F$ is the number of times the PDA {\em repeated} step 5.1 (i.e. executed a ``fallback-again" rule of 5.2), and $M$ is the number of times it ends the loop of 5), i.e. selects a non-``fallback-again" rule. Each step $i$ will correspond to a ``strong"-step of an equivalent FBA execution, which is a transition where the input symbol has type not equal to $s$ (in other words, a step where the FBA either processes a fresh symbol, or falls back to a marked symbol).

Claim: There is a one-to-one correspondence between executions of the FBA on $x$ with executions of the PDA on $x$, such that, for each such pair, the PDA simulation maintains the following invariant at each step $i$ with respect to the FBA:

a) the PDA state {\em after} the $i$-th step is equal to the FBA state immediately before the $i+1$-th strong step (or the final state, if $i=|x|$), and the position of the PDA head is equal to the position of the last $f$-symbol seen by the FBA.

b) the PDA stack has $l$ elements, where $l$ is the number of marked symbols after the $i$-th strong-step of the FBA, and 

c) the $k$-th stack element consists of the $k$-th marked symbol and type in the FBA execution, and a vector that contains for each state $q_j$, the execution of $s$-transitions starting from the symbol immediately after the $k$-th marked symbol, up to and including either the $(k+1)$-th marked symbol, or the head, whichever comes first.

Note that if the claim is true, by part a) of the invariant, the final state of the PDA is an accept state iff the equivalent FBA execution accepts $x$, so the theorem is proved.

We prove the claim recursively. It is trivially true at the beginning. Now, assume that after $i-1$ steps of execution, the FBA corresponds to an execution of $i-1$ strong-steps of the PDA, and that the invariant is true.

At step $i$, we consider all possible strong-steps of the FBA.

First, for any rule of the type $((\alpha,f,q'),(q,s)) \in \Delta$: by a) of the invariant the FBA and PDA must both be scanning a new symbol $\alpha$ and are in state $q'$. The PDA simulation can sample this rule by step 1), which results in updating the state $q'$ to $q$, satisfying a). Since no marked symbols were added or removed by this kind of FBA transition, b) is trivially satisfied. Finally, 2) updates the top state vector on the stack with the $s$-transitions on $\alpha$-- since it previously represented the $s$-transitions from the most recent marked symbol to the previous tape symbol, it now represents an execution up to and including the current symbol, that is, c) is maintained.

For any rule of the type $((\alpha,f,q'),(q,m)) \in \Delta$: similarly, the PDA simulation can select this rule in 1), updating the state to $q'$, giving a). The number of marked symbols in the FBA execution changes if this rule were applied, so invariants b) and c) must be checked. By step 3), we push $(\alpha, q, (q_1,\ldots,q_t))$ to the stack, immediately satisfying b). Note that the penultimate element of the stack is now ``frozen", showing an execution of $s$-transitions from the previous marked-symbol up to and including $\alpha$, the new marked symbol. The vector of the top element of the stack, $(q_1,\ldots,q_t)$, trivially represents an execution of every state from the new marked symbol to the head (since it is empty). Hence c) is satisfied.

For any rule of the type $((\alpha,f,q'),(q,\leftarrow)) \in \Delta$: again both the PDA and FBA are at a fresh tape symbol, but after this step, the FBA head will be at the previously marked symbol. By step 1) the PDA can sample this rule, updating the state to $q$, which satisfies a). Note that b) and c) are trivially satisfied.

Next for any strong rule of the type $((\beta,p,q'),(q,\leftarrow))$, the FBA head must be at a marked symbol, and immediately before the next strong step, it will be at the previous marked symbol in state $q$. Indeed, the FBA can sample this rule in 5.1, updates state to $q$ satisfying a), pops the top stack vector satisfying b), and as for c),  the PDA ``applies" the state vector of the popped tuple, $(r_1,\ldots,r_t)$, to the {\em next} state vector on the stack, $(\gamma,w,(u_1,\ldots,u_t))$. If $u_i$ is the $s$-transition of $q_i$ from the symbol after $\gamma$ up to and including $\beta$ (this is guaranteed by the invariant) and $u_i = q_j$, then $r_j$ is exactly the $s$-transition of $q_i$ from the symbol after $\beta$ up to and including the head, ensuring c).

Finally if the rule is of the type $((\beta,p,q'),(q,s)$, the FBA head must be at a marked symbol, and immediately before the next strong step it {\em passes through every s-symbol between the marked symbol and the fresh symbol}. The corresponding PDA transition satisfies b) and c) as in the previous case (a marked symbol is removed and the other vectors in the stack are updated), but this time, we also update the current state based on the popped vector; since, by b), it contained the $s$-transition of each state from the marked symbol to the head, it correctly yields the state of the FBA before the next fresh symbol.

\bibliography{parser, anthology}
\bibliographystyle{acl_natbib}

\end{document}